\documentclass[lettersize,journal]{IEEEtran}
\usepackage{amsmath,amsfonts}
\usepackage{algorithmic}
\usepackage{algorithm}
\usepackage{array}
\usepackage[caption=false,font=normalsize,labelfont=sf,textfont=sf]{subfig}
\usepackage{textcomp}
\usepackage{stfloats}
\usepackage{url}
\usepackage{verbatim}
\usepackage{graphicx}
\usepackage{cite}
\usepackage{multirow}
\usepackage{booktabs}
\usepackage{caption}
\usepackage{subcaption}
\usepackage{makecell}
\usepackage{subfig}
\usepackage[normalem]{ulem}
\useunder{\uline}{\ul}{}
\usepackage[table,xcdraw]{xcolor}

\begin{document}

\title{DocShaDiffusion: Diffusion Model in Latent Space for Document Image Shadow Removal}

\author{Wenjie Liu, Bingshu Wang*, Ze Wang, C.L. Philip Chen 
\thanks{This work is supported by the National Natural Science Foundation of China, Youth Fund, under number 62102318, and the Basic Research Programs of Taicang, 2023, under number TC2023JC23. This work is funded in part by the National Natural Science Foundation of China grant under number 92267203, in part by the Science and Technology Major Project of Guangzhou under number 202007030006, and in part by the Program for Guangdong Introducing Innovative and Entrepreneurial Teams (2019ZT08X214). (Corresponding author: Bingshu Wang) 

Wenjie Liu is with the School of Software, Northwestern Polytechnical University, Xi’an 710129, China (e-mail: xjie@mail.nwpu.edu.cn). 

Bingshu Wang is with the School of Software, Northwestern Polytechnical University, Xi’an 710129, China (e-mail: wangbingshu@nwpu.edu.cn).

Ze Wang is with the School of Software, Northwestern Polytechnical University, Xi’an 710129, China (e-mail: wangze880@mail.nwpu.edu.cn). 
 
C. L. P. Chen is with the School of Computer Science and Engineering,
South China University of Technology and Pazhou Lab, Guangzhou 510641, 510335, China (e-mail: philip.chen@ieee.org).}
\thanks{}}

\markboth{Journal of \LaTeX\ Class Files,~Vol.~14, No.~8, August~2021}%
{Shell \MakeLowercase{\textit{et al.}}: DocShaDiffusion: Diffusion Model in Latent Space for Document Image Shadow Removal}

\IEEEpubid{}

\maketitle

\begin{abstract}
Document shadow removal is a crucial task in the field of document image enhancement. However, existing methods tend to remove shadows with constant color background and ignore color shadows. In this paper, we first design a diffusion model in latent space for document image shadow removal, called DocShaDiffusion. It translates shadow images from pixel space to latent space, enabling the model to more easily capture essential features. To address the issue of color shadows, we design a shadow soft-mask generation module (SSGM). It is able to produce accurate shadow mask and add noise into shadow regions specially. Guided by the shadow mask, a shadow mask-aware guided diffusion module (SMGDM) is proposed to remove shadows from document images by supervising the diffusion and denoising process. We also propose a shadow-robust perceptual feature loss to preserve details and structures in document images. Moreover, we develop a large-scale synthetic document color shadow removal dataset (SDCSRD). It simulates the distribution of realistic color shadows and provides powerful supports for the training of models. Experiments on three public datasets validate the proposed method's superiority over state-of-the-art.  Our code and dataset will be publicly available. 
\end{abstract}


\section{Introduction}
\IEEEPARstart{W}{ith} the widespread use of electronic devices such as smartphones, people are increasingly inclined to use document images for information storage \cite{55mondal2022statistical,56georgiadis2023lp,57bogdan2021ddoce}. However, shadows \cite{tmm1, tmm2} in document images has consistently posed a challenge \cite{43li2019document, 44he2019deepotsu, 45huang2020binarization, 53liu2023docstormer}, reducing the readability and quality of document images. As a result, this impedes tasks such as information extraction, digital archive management, and automated processing \cite{58dey2021light,60li2021document, 61zhang2023appearance}. Therefore, the removal of shadows from document images is an important task for vision applications \cite{59imahayashi2023shadow, 50shadow2022udocgan,52luo2023docdeshadower}.

\begin{figure}[h]
	\centering
	\includegraphics[width=\linewidth]{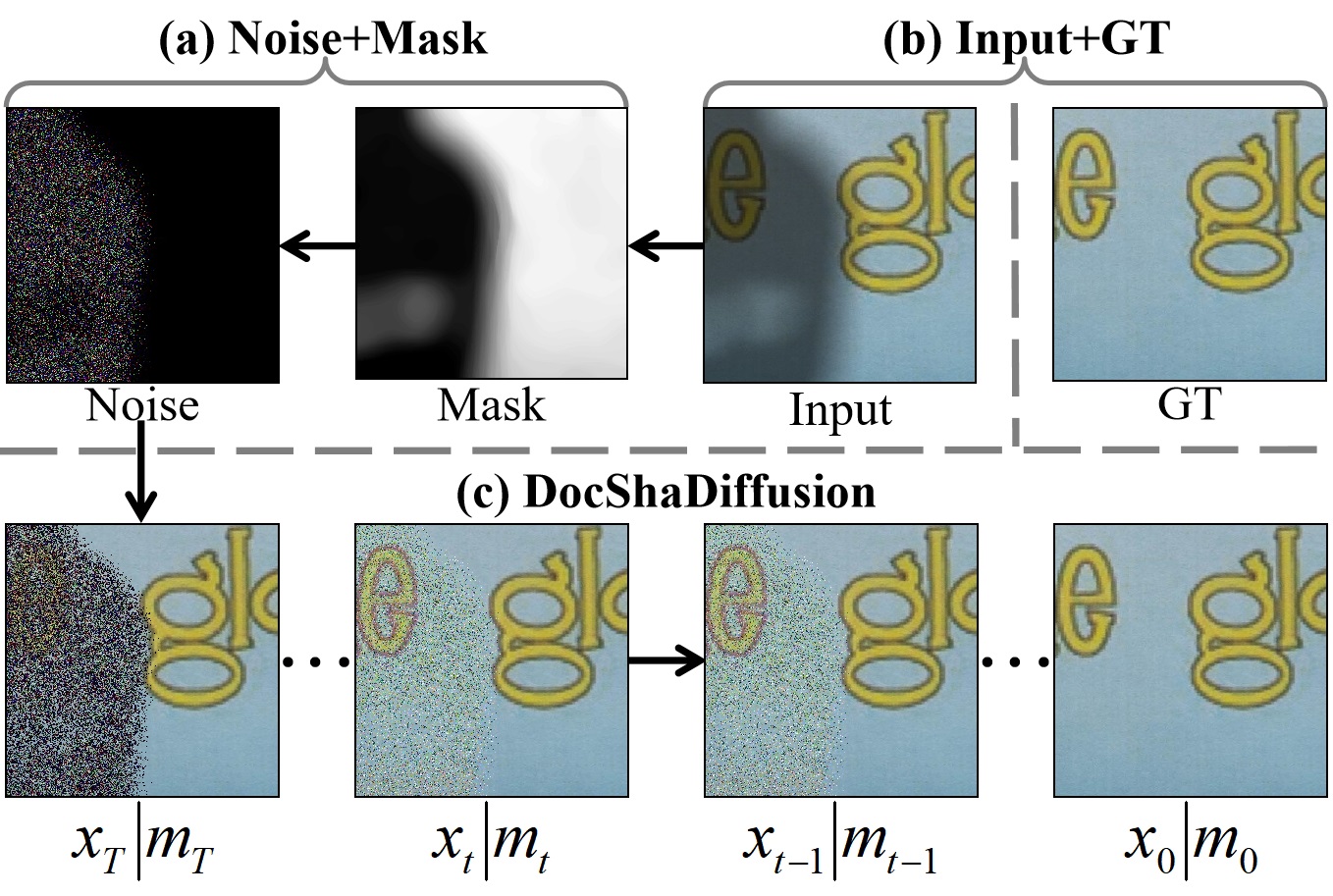}  
   \caption{An example of our proposed DocShaDiffusion for document image shadow removal. (a) noise and mask of our DocShaDiffusion, (b) input shadow image and corresponding ground truth, (c) our proposed DocShaDiffusion iteratively ($T\rightarrow0$) restores the shadow-free image. } 
	\label{fig:objective}
\end{figure}

Researchers around the world have proposed many methods to tackle the task, which can be categorized into two groups: heuristic-based methods \cite{39Bako2016, 01wangze2022, 02liuwenjie2023, 26kligler2018comparative, 27jung2018comparative} and neural network-based methods \cite{51anvari2021survey, 34unsupervised2020nature, 36via2019nature, 37self2020nature}. Heuristic-based methods are often tailored to specific scenarios. Kligler et al. \cite{26kligler2018comparative} proposed a 3D point cloud transformation method for visual detection. Jung et al. \cite{27jung2018comparative} proposed a water-filling method inspired by the immersion process of topographic surface with water. However, these methods have difficulties in handling complex scenarios with different types of shadows.

Neural network-based methods \cite{11macau2023icassp, 12macau2024icassparXiv, 13macau2023ICCV, 42jie2022fast} have been proposed with generalization abilities over datasets. Lin et al. \cite{41lin2020bedsr} introduced the first neural network framework for shadow removal from document images, BEDSR-Net. Zhang et al. \cite{10RDD2023cvpr} proposed CBENet to estimate spatially varying backgrounds and retain more background details, and used BGShadowNet for shadow removal, but this approach tends to produce noise in heavy shadows. These methods can obtain good results on soft shadows, however, when facing color shadows, they may produce artifacts.

The motivation of this paper is to design a neural network to handle multiple scenes. We propose a diffusion model called DocShaDiffusion. It facilitates noise diffusion in the latent space with the assistance of shadow soft-mask, as shown in Fig.\ref{fig:objective}.

Moreover, the training of neural network-based models requires a large scale of images. Matsuo et al. \cite{47matsuo2022fsdsrd} proposed a synthetic dataset FSDSRD, which includes 14200 pairs of black shadow images. However, it does not properly simulate the real shadow distribution. In this paper, we propose a complex synthetic shadow dataset containing 17624 pairs of color shadows, as shown in Fig.\ref{fig:dataset}.

Our contributions are presented as follows: 

\begin{itemize}
    \item{We propose the first diffusion-based model called DocShaDiffusion for document image shadow removal. It consists of forward noise addition and reverse denoising in latent space to learn robust feature maps. }
    \item{We design a shadow soft-mask generation module (SSGM). It aims to generate shadow soft-mask, which is used to add random noises into specific shadow regions. }
    \item{We design a shadow mask-aware guided diffusion module (SMGDM). By supervising the diffusion and denoising process with a shadow mask-aware mechanism, it can selectively address shadowed regions while preserving the detail and quality of non-shadowed areas. }
    \item{We propose a synthetic document color shadow removal dataset (SDCSRD), which simulates the real distribution of shadows. Experiments on SDCSRD and other datasets show that the proposed DocShaDiffusion outperforms the state-of-the-art (SOTA) methods. }
\end{itemize}


\section{Related work}

\subsection{Document Image Shadow Removal Datasets}

Existing document image shadow removal datasets can be categorized into two types: real datasets \cite{39Bako2016, 27jung2018comparative, 26kligler2018comparative, 25iterative2018comparative, 46wang2020dataset, 10RDD2023cvpr} and synthetic datasets \cite{47matsuo2022fsdsrd}. Adobe dataset \cite{39Bako2016} is the first dedicated to the task of shadow removal in document images, containing only handwritten texts. RDD dataset \cite{10RDD2023cvpr} is a large real-world dataset for document image shadow removal, consisting of 4914 pairs of shadow and shadow-free images. Capturing document images in the real world presents challenges due to environmental factors such as lighting, shadow intensity, and background. Thus, it is difficult to control environmental factors such as lighting and shadow intensities in document images. 

Simulating document images with shadows is considered as a good way to create datasets. One typical dataset is FSDSRD \cite{47matsuo2022fsdsrd}, but it is far from the distribution of real shadows, as shown in Fig.\ref{fig:datasetHsv}. To simulate real document images with shadows, we create a dataset called SDCSRD by considering natural illuminations and shadows distribution.

\subsection{Shadow Removal Methods}

Earlier document image shadow removal methods are primarily heuristic-based methods \cite{02liuwenjie2023, 62jiang2015illumination, 26kligler2018comparative, 01wangze2022, 29wang2020comparative}. They can be divided into two classes: illumination model-based methods \cite{62jiang2015illumination, 26kligler2018comparative} and shadow map-based methods  \cite{01wangze2022, 29wang2020comparative, 02liuwenjie2023}. For illumination model-based methods, Kligler et al. \cite{26kligler2018comparative} introduced a 3D point cloud transformation method based on visual detection, which generates a new image representation from the original image. However, it easily leads to artifacts in  penumbra. For shadow map-based methods, Jung et al. \cite{27jung2018comparative} and  Wang et al.\cite{46wang2020dataset} considered to use water filling algorithm to simulate the submersion process of surface water. Nonetheless, it easily produces  color artifacts in shadow regions. In summary, the heuristic-based methods are usually designed for specific scenarios, and their adaptation abilities over scenes are limited by the heuristic design.

Neural network-based methods \cite{11macau2023icassp, 12macau2024icassparXiv, 13macau2023ICCV, 42jie2022fast, 41lin2020bedsr, 47matsuo2022fsdsrd, 10RDD2023cvpr} are usually designed with multiple stages, to learn features from images automatically. Lin et al. \cite{41lin2020bedsr} introduced BEDSR-Net, the first neural network-based method for document shadow removal. However, it only considers the situation of  constant color for images.  Matsuo et al. \cite{47matsuo2022fsdsrd} proposed a  tansformer-based method for shadow removal. It uses the outputs from BEDSR-Net as the inputs of tansformer-based model, which adds its complexity.
  
In summary, due to the effectiveness of multiple stages of feature extraction, Neural network-based methods outperforms heuristic-based methods. Thus, this paper designs a two-stage method by firstly introducing diffusion model into the field of document image shadow removal.

\subsection{Diffusion Model}

Recently, diffusion-based generative models \cite{05docdiff2023arXiv, 06cvpr2023shadowdiffusion, 63cross2023diffusion} have produced astonishing results. These models create samples by inversely simulating the iterative process of adding noise to data, learning to recover clean data from noise. The denoising diffusion probabilistic models \cite{16ddpm2020, 17iddpm2021} and denoising diffusion implicit models \cite{18ddim2020} are increasingly influencing the field of low-level vision tasks, such as super-resolution \cite{19superresolution2022diffusion, 20superresolution2022diffusioniterative}, inpainting \cite{21inpainting2022diffusion}, coloring \cite{22colorization2022diffusion}, shadow removal \cite{06cvpr2023shadowdiffusion}, document enhancement \cite{05docdiff2023arXiv,49Binarization2023Diffusion}, rain removal \cite{23raindiffusion2023}, and fog removal \cite{24dehazingiffusion2023}. Yang et al. \cite{05docdiff2023arXiv} introduced Docdiff, which employs residual diffusion models for document image enhancement. Guo et al. \cite{06cvpr2023shadowdiffusion} applied diffusion model into natural shadow removal.

Although diffusion models have achieved impressive results in many scenes, it is hard to transform them into document image shadow removal directly. This is primarily due to the inherent differences between natural scenes and document images, such as variations in lighting conditions and shadow morphologies. Thus, it is required to design a diffusion model for document image shadow removal. 


\section{The Proposed Dataset}

\subsection{How to Create the Dataset?}

\begin{figure}[h]
	\centering
	\includegraphics[width=\linewidth]{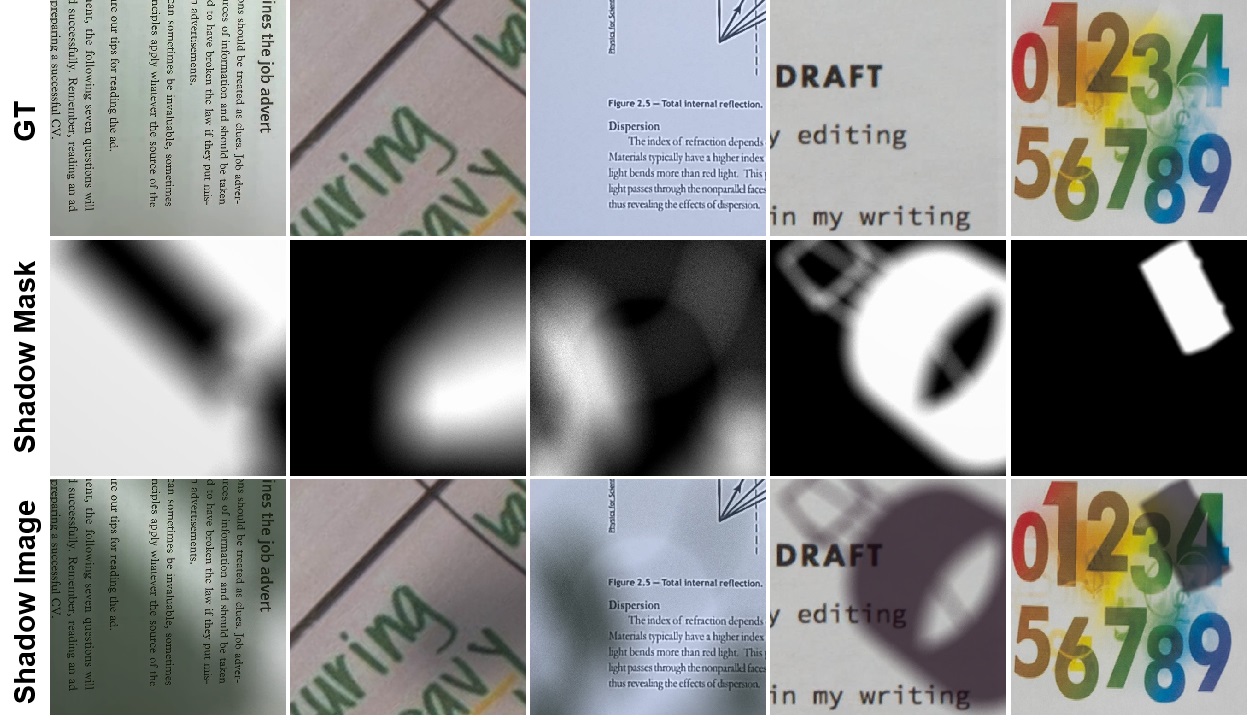}  
   \caption{Examples of our dataset (SDCSRD). From up to down: ground truth, shadow mask, and shadow image. } 
	\label{fig:dataset}
\end{figure}

We propose a synthetic document color shadow removal dataset (SDCSRD), as shown in Fig.\ref{fig:dataset}. First, we collect shadow-free images (ground truth) from the real dataset. We process these images through rotation, magnification, translation, and other operations, cropping them into $ 512 \times 512$ pixel images. 

Then, we obtain the shadow masks from Inoue et al. \cite{03inoue2020dataset} and FSDSRD \cite{47matsuo2022fsdsrd}, treating them as shadow mask templates. We first perform histogram equalization on the mask template to ensure that each soft-mask has non-shadow regions set to 0 and shadow regions transitioning $(0,1]$, with 1 representing the darkest area of the shadow. Subsequently, we apply random scaling and rotation to the equalized dataset and crop them to the final $ 512 \times 512$ shadow soft-masks, as shown in the second row of Fig.\ref{fig:dataset}.

Here, $\mathfrak{a}$ represents the weight of the shadow, a random value between 0 and 1, and ${C}_{k}$ is a randomly generated shadow color. We use the Eq.\ref{con:ourdatasetEq} to generate a colored shadow dataset.

\begin{equation}
  \begin{aligned} 
    {I}_{k}^{s} = {\mathfrak{a}}{I}_{k}^{sf} + \left ( {1-\mathfrak{a}} \right ) {I}_{k}^{m}{C}_{k} \label{con:ourdatasetEq}\\[0.2mm]
    \end{aligned}
 \end{equation} 
 where both ${\mathfrak{a}}$ and ${C}_{k}$ are scalar values. ${I}^{s}$, ${I}^{sf}$ and ${I}^{m}$ represent the shadow-free image, shadow image and shadow mask in the RGB color space, respectively.

\begin{figure}[h]
	\centering
	\includegraphics[width=\linewidth]{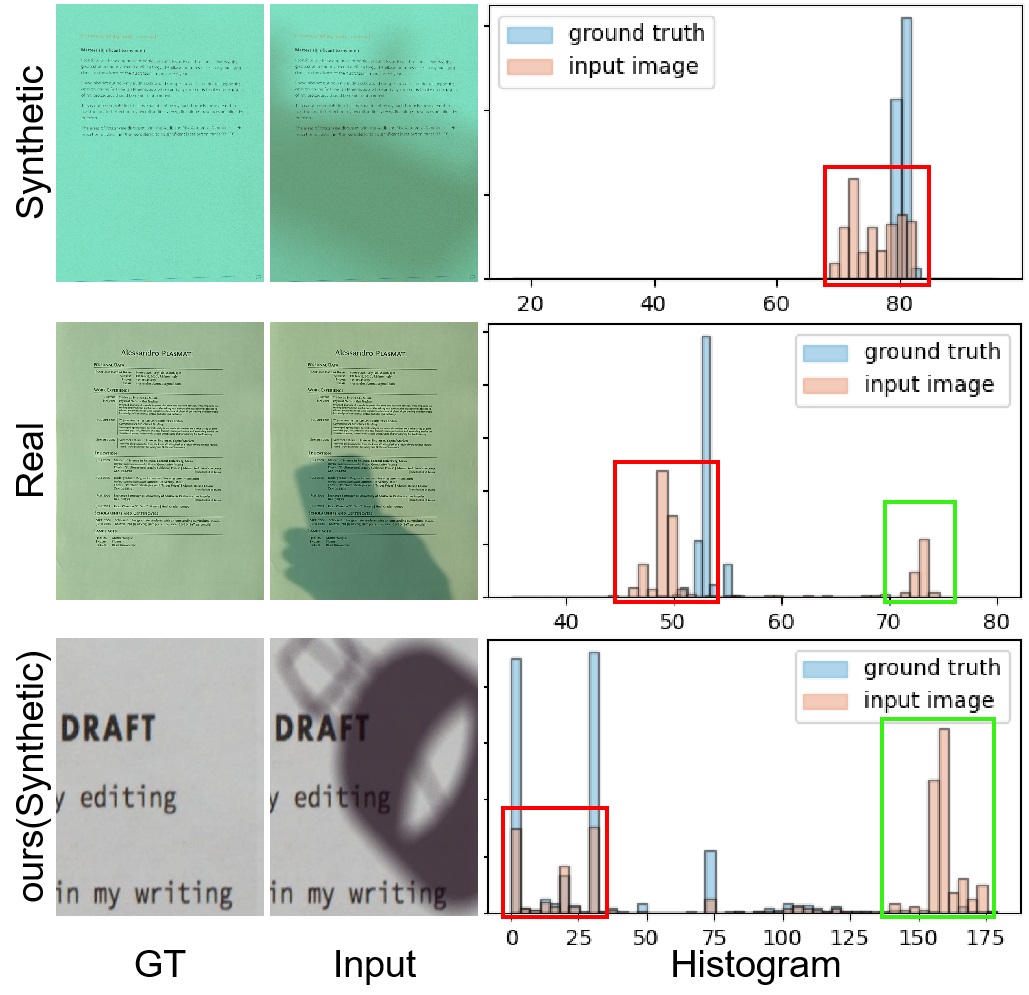}  
   \caption{Comparison of color distribution histograms between existing synthetic datset (FSDSRD), real dataset (HS) and our synthetic datset (SDCSRD), respectively.} 
	\label{fig:datasetHsv}
\end{figure}

\subsection{The Characteristics of the Dataset}

\begin{figure*}[h]  
	\makebox[\textwidth][c]{\includegraphics[width=7in]{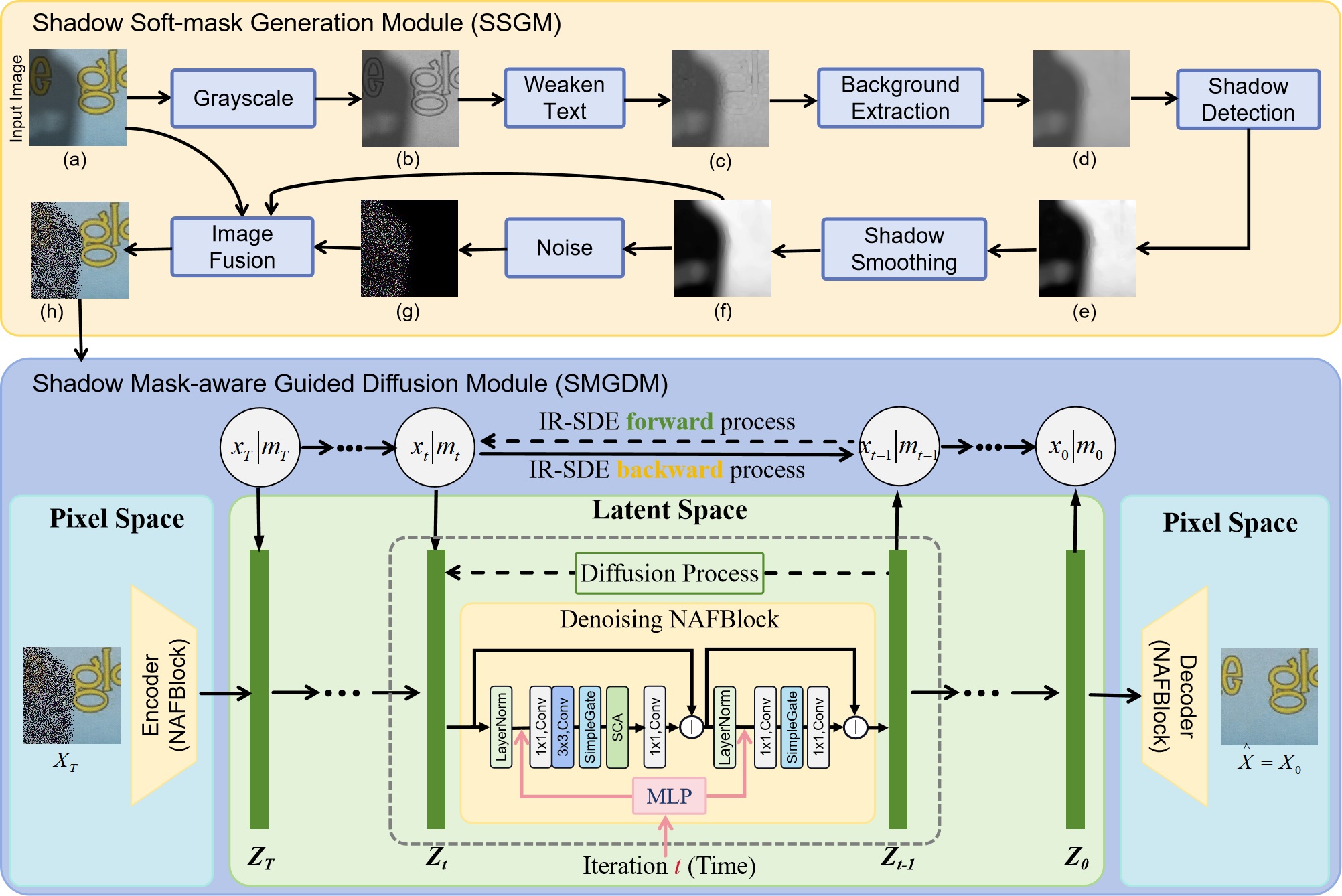}}  
	\caption{Illustration of the proposed DocShaDiffusion, in which the training process is dashed line and sampling process is solid line. $X_T$ is shadow image with noise, $X_0$ is shadow-free image, $Z_{T-0}$ is the feature map in latent space of $X_{T-0}$ each step.  }  
	\label{fig:flowchart}
\end{figure*}


It can be observed from Fig.\ref{fig:datasetHsv} that the distribution of our color shadow dataset closely resembles that of the real dataset. The second row shows the color distribution of a real dataset, and we can observe that shadows have altered the background color distribution of document images, as indicated by the green box. However, the synthetic FSDSRD in the first row does not have the discriminative background distribution. In contrast, our synthetic dataset, as illustrated in the third row, is similar with the background distribution of HS. This is demonstrated by the green boxes in Fig.\ref{fig:datasetHsv}. 

Moreover, in terms of the number of images for datasets, the number of images in our SDCSRD dataset is more than FSDSRD. Our dataset has a total of 17624 images, they are divided into three subsets at a ratio of 12:3:1 for train, valid, and test sets, respectively. This is expected to provide powerful support for model training of deep learning. 


\section{Method}

In this section, we introduce the proposed DocShaDiffusion, as shown in Fig.\ref{fig:flowchart}. It consists of two modules: shadow soft-mask generation module (SSGM) and shadow mask-aware guided diffusion module (SMGDM). SSGM is designed to obtain shadow soft mask from a shadow image. Subsequently, SMGDM incrementally predicts the noise distribution within the shadow regions to ultimately yield a shadow-free image. Moreover, considering that training a diffusion model in pixel space is slow and yields suboptimal results, we initially employ an encoder to transform the input image into a feature map in the latent space for diffusion. Finally, we give the loss function design. 

\subsection{Shadow Soft-mask Generation Module}

Existing diffusion models are designed by generating the target image from an initial pure noise representation. Unlike this, in the task of document image shadow removal, our focus is exclusively on the shadow regions \cite{14HuXiaowei2021Dataset, 14HuXiaowei2022shadowdetection, 35distraction2019nature, 38multi2020nature}. By directing the model’s attention to the shadows rather than the entire image, it not only accelerates the training process, but also enables the trained model to learn the shadow information more effectively. Therefore, we propose the shadow soft-mask generation module (SSGM) to realize the function, as shown in Fig.\ref{fig:flowchart}.

To enhance brightness discrepancies and aid in shadow contour detection, converting color images to grayscale is crucial, as shown in Fig.\ref{fig:flowchart}(b). Additionally, dilation operations refine this process by removing elements like text and noise that could be misinterpreted as shadows, as depicted in Fig.\ref{fig:flowchart}(c). Finally, we obtain the background image by median filtering to smooth the image, effectively eliminating random noise while still emphasizing the shadow edges, maintaining crisp edge detail as illustrated in Fig.\ref{fig:flowchart}(d).

Then, we extract the shadows based on the obtained background image. Adjusting shadow brightness using Eq.\ref{con:numEq} and Eq.\ref{con:meanEq}, followed by normalization with Eq.\ref{con:ImEq}, is vital for crafting a preliminary mask that accurately reflects shadow luminance, as illustrated in Fig.\ref{fig:flowchart}(e). This process not only aids in precise shadow region identification but also compensates for luminance disparities due to uneven lighting. A subsequent round of median filtering, shown in Fig.\ref{fig:flowchart}(f), then further refines the mask by smoothing and noise reduction, resulting in a shadow soft-mask that faithfully represents the shadowed areas in the original image, providing a high-quality basis for noise addition.

\begin{equation}
  \begin{aligned} 
    m = a \cdotp ({w \times h}), a\in (0,1) \label{con:numEq}\\[0.2mm]
  \end{aligned}
\end{equation} 
 where $m$ represents the number of pixels selected, and $w$ and $h$ represent the image's width and height, respectively.

 \begin{equation}
  \begin{aligned} 
    {p}_{mean} = \frac{1}{m}\displaystyle\sum_{i=0}^{m-1}{I}_{sort}(i) \label{con:meanEq}\\[0.2mm]
    \end{aligned}
 \end{equation} 
 where ${p}_{mean}$ represents the mean of the selected in the image. ${I}_{sort}$ represents the vector sorted by image pixel values from smallest to largest. 

 \begin{equation}
  \begin{aligned} 
    {I}^{m} = \frac{{I}^{s}-{p}_{mean}}{{p}_{max}-{p}_{mean}}\label{con:ImEq}\\[0.2mm]
    \end{aligned}
 \end{equation} 
 where ${p}_{mean}, {p}_{max}$ represent the mean of the selected and max pixel in the image, respectively. ${I}^{s}, {I}^{m}$ represent shadow image and shadow mask, respectively.  

\begin{algorithm}
\caption{Generate shadow soft-mask} \label{alg:mask}
\begin{algorithmic}[1]
\renewcommand{\algorithmicrequire}{\textbf{Input:}}
 \renewcommand{\algorithmicensure}{\textbf{Output:}}
\REQUIRE shadow image ${I}^{s}$
\ENSURE shadow soft-mask ${I}^{m}$
\STATE ${I}_{gray} \gets $ RGB2Gray(${I}^{s}$)
\STATE ${I}_{dilated} \gets $ Dilate(${I}_{gray}$)
\STATE ${I}_{filtered} \gets $ MedianFilter(${I}_{dilated}$)
\STATE ${I}_{sort} \gets $ Sort(${I}_{filtered}$) 
\STATE ${I}_{mask} \gets $ GenerateMask(${I}_{filtered}$, ${I}_{sort}$) 
\STATE ${I}^{m} \gets $ MedianFilter(${I}_{mask}$)
\RETURN ${I}^{m}$
\end{algorithmic}
\end{algorithm}

 Based on the shadow soft-mask obtained from Algorithm \ref{alg:mask}, we can supervise the intensity of noise when generating a pure gaussian noise image using the normalized shadow soft-mask, thereby producing a shadow mask-aware guided noise image, as shown in Fig.\ref{fig:flowchart}(g). Subsequently, we merge the input image with the noise image to obtain $X_T$, as illustrated in Fig.\ref{fig:flowchart}(h).

\subsection{Shadow Mask-aware Guided Diffusion Module}

Through SSGM, we obtain the corresponding shadow soft-mask for each input image, as shown in Fig.\ref{fig:flowchart}. With the guidance of the shadow soft-mask, our diffusion model can concentrate on processing the shadow regions without exerting unnecessary influence on non-shadow areas. 


Our method utilizes IR-SDE \cite{07refusion2023latent} as the foundational diffusion framework, capable of naturally transitioning images with shadows into high-quality shadow-free images (as shown in Fig.\ref{fig:flowchart}). The forward process of IR-SDE is defined as follows Eq.\ref{con:forwardEq}:

\begin{equation}
  \begin{aligned} 
    dx = {\theta }_{t}(\mu -x)dt-{\sigma }_{t}(1-{I}^{m})dw \label{con:forwardEq}\\[0.2mm]
    \end{aligned}
 \end{equation} 
 where ${\theta }_{t}$ and ${\sigma }_{t}$ are time-related parameters, describing the average reversion rate and random volatility, respectively. $I^m$ is the shadow mask got from Algorithm \ref{alg:mask}. "$x$" signifies an image with shadows, while $\mu$ represents the target image without shadows. $dw$ represents noise.
 
 The randomness in this process originates from the complexity and uncertainty of factors such as lighting conditions and surface reflectivity of objects in the real world. Its reverse time representation is given by Eq.\ref{con:backwardEq}. 

\begin{equation}
  \begin{aligned} 
    dx = [{\theta }_{t}(\mu -x)-{\sigma }_{t}^{2}{\nabla }_{x}\log{{p}_{t}(x)}]dt-{\sigma }_{t}(1-{I}^{m})d\hat{w} \label{con:backwardEq}\\[0.2mm]
    \end{aligned}
 \end{equation} 
 where ${\nabla }_{x}\log{{p}_{t}(x)}$ represents the score function of $x$ at time $t$. The noise term, with the added supervision of the shadow-mask, implies that the impact of random disturbances is amplified in shadow areas. Conversely, in shadow-free regions, the influence of random noise disturbances is diminished.

The experimental results indicate that executing the denoising process in pixel space would necessitate excessive computational resources. We compress the entire process into a latent space to reduce computational resources. A pre-trained UNet network is introduced, where in the shadow image is first compressed into a latent space through an encoder, then converted into a shadow-free image's latent representation via the reverse denoising process of the diffusion model \cite{07refusion2023latent, 15latentdiffusion2022}. Subsequently, a shadow-free image is reconstructed from this by a decoder.

We employ a NAFBlock\cite{04nafblockchen2022, 07refusion2023latent} with time encoding to replace the encoder and decoder of the UNet. As shown in Fig.\ref{fig:flowchart}, the temporal information is processed via a Multi-Layer Perceptron (MLP). Specifically, the time encoding undergoes a sine position embedding layer, and then, transformed by two linear layers, it ultimately obtains a new time embedding $t$. During the forward propagation process, this new time embedding $t$ is passed to each NAFBlock layer, allowing these layers to take into account temporal information. In particular, in each NAFBlock, the time embedding $t$ is utilized to change the scaling and offset of the feature map, thereby controlling the transformation mode of the feature map, achieving effects akin to those of an encoder and decoder in a UNet architecture.

\subsection{Loss Function}


The loss function $L_{total}$ comprises two parts, the diffusion objective loss and shadow-robust feature loss, denoted as $L_{diff}$ and $L_{fea}$ respectively.

The diffusion objective loss function is as Eq.\ref{con:LdiffEq}.

\begin{equation}
  \begin{aligned} 
    {\mathcal{L}}_{diff} = {\mathbb{E}}_{{x}_{0},t,\epsilon}{\parallel{e}_{t}-\epsilon  \parallel }_{F}^{1} \label{con:LdiffEq}\\[0.2mm]
    \end{aligned}
 \end{equation} 
 where ${e}_{t}$ represents the predicted noise map. $\epsilon$ represents the noise of input image.  

 \begin{figure}[h]
   \centering
   \includegraphics[width=\linewidth]{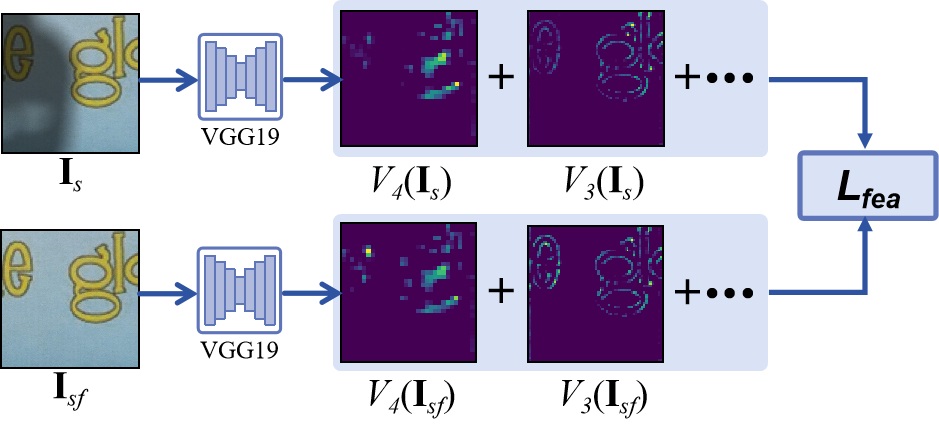}
   \caption{Illustration of the proposed ${L}_{fea}$. $I_{s}$ is shadow image. $I_{sf}$ is shadow-free image. $V_{i}$ is the feature map in VGG-19.  }  
   \label{fig:loss}
\end{figure}

 \begin{figure*}[h]
	\centering
	\includegraphics[width=\linewidth]{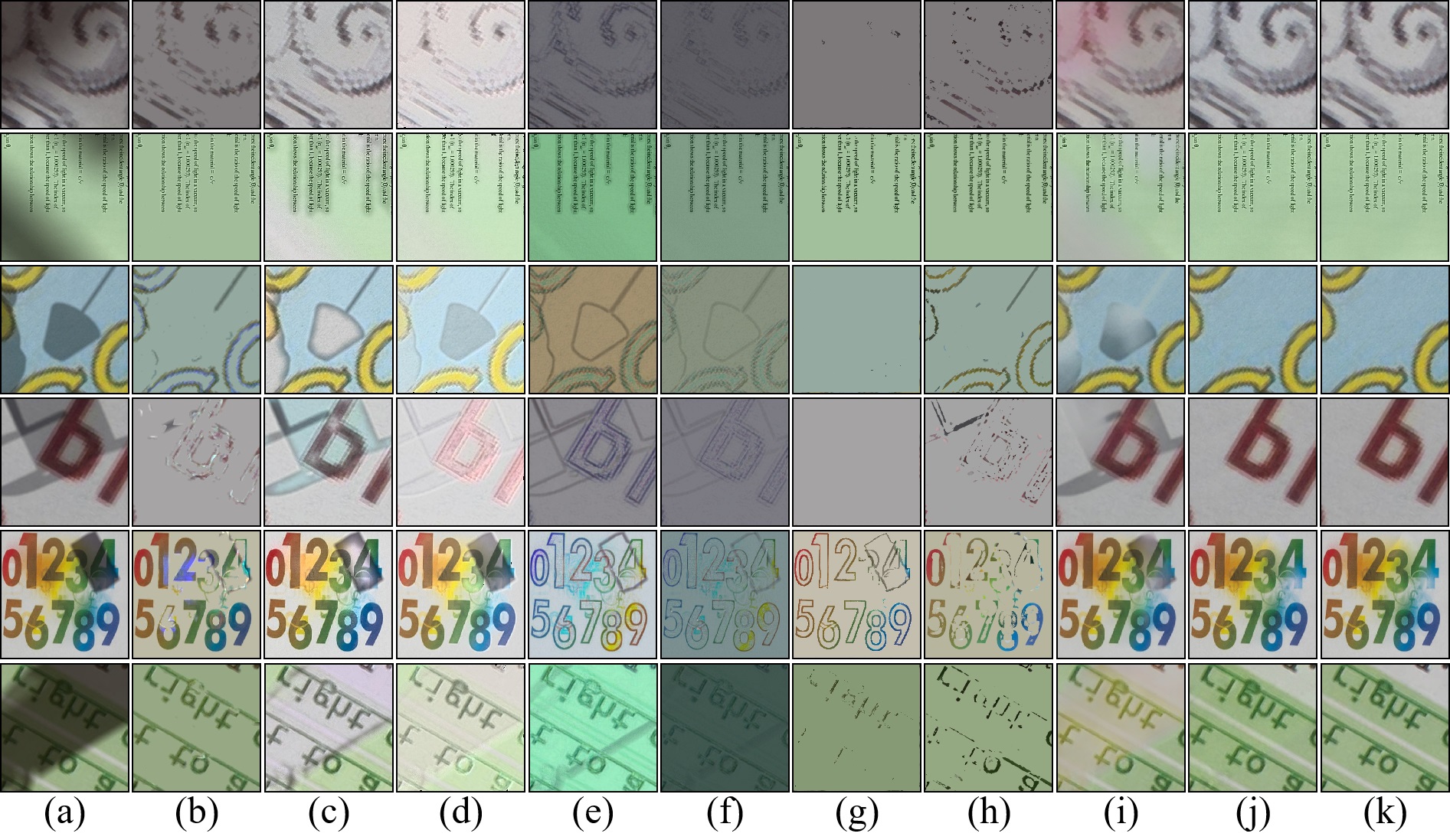}  
   \caption{Visual comparison between our method and the SOTA methods. (a)input images, (b)ISR-based\cite{25iterative2018comparative}, (c)3D-PC-based\cite{26kligler2018comparative}, (d)WF-based\cite{27jung2018comparative}, (e)BE-based\cite{28wang2019comparative}, (f)LGBC-based\cite{29wang2020comparative}, (g)JWF-based\cite{01wangze2022}, (h)BEATE-based\cite{02liuwenjie2023}, (i)CBENet\cite{10RDD2023cvpr},   (j)ours, (k)gt. } 
	\label{fig:VisualResults}
\end{figure*}

\begin{table*}[ht]
  \begin{center}
  \begin{tabular}{*{8}{c|c|c|ccc|ccc}}
  \toprule
  & \multirow{2}{*}{\makecell[c]{Methods}} &  \multirow{2}{*}{\makecell[c]{Venue/Year}} & \multicolumn{3}{c|}{\makecell[c]{Adobe Dataset}} & \multicolumn{3}{c}{\makecell[c]{FSDSRD Dataset}}  \\
  & & & \makecell[c]{$PSNR\uparrow$} &  \makecell[c]{$SSIM\uparrow$} &  \makecell[c]{$RMSE\downarrow $} & \makecell[c]{$PSNR\uparrow$} &  \makecell[c]{$SSIM\uparrow$} &  \makecell[c]{$RMSE\downarrow $}  \\
  \midrule
  \multirow{7}{*}{\rotatebox{90}{\parbox[c]{3cm}{\centering Heuristic}}} & ISR-based\cite{25iterative2018comparative} & ICASSP/2018 & 29.83 & {\color[HTML]{0045FE} \textbf{ 0.95}} & 8.55   & 26.57 & 0.91 & 15.24 \\
  &3D-PC-based\cite{26kligler2018comparative} & CVPR/2018 & 15.06 & 0.79 & 45.11  & 24.62 & 0.92 & 17.15 \\
  & WF-based\cite{27jung2018comparative} & ACCV/2018 & 10.23 & 0.77 & 79.11  & 21.30 & {\color[HTML]{3166FF} \textbf{ 0.96}} & 25.14 \\
  & BE-based\cite{28wang2019comparative} & ICIP/2019 & 13.65 & 0.71 & 55.00  & 1.22 & $\approx $0 & 222.63 \\
  & LGBC-based\cite{29wang2020comparative} & ICASSP/2020 & 21.34 & 0.91 & 23.77  & 16.72 & 0.93 & 44.08 \\
  & JWF-based\cite{01wangze2022} & SMC/2022 & 24.72 & 0.85 & 15.62 & 26.00 & 0.85 & 16.68 \\
  & BEATE-based\cite{02liuwenjie2023} & ICASSP/2023 & 27.95 & 0.91 & 10.46 & 24.22 & 0.83 & 19.00 \\
  \midrule
  \multirow{4}{*}{\rotatebox{90}{\parbox[c]{1.7cm}{\centering Neural Network}}} &  MS-GAN\cite{30maskgan2019comparative} & CVPR/2019 & 28.30 & 0.85  & 10.37  & 25.42 & 0.80 & {\color[HTML]{0045FE} \textbf{14.73}} \\
  & DCShadow-Net\cite{32Dcshadownet2021comparative} & ICCV/2021 & {\color[HTML]{0045FE} \textbf{ 32.25}} & {\color[HTML]{0045FE} \textbf{ 0.95}} & {\color[HTML]{0045FE} \textbf{ 6.51}}  & {\color[HTML]{0045FE} \textbf{ 26.83}} & 0.89 & {\color[HTML]{FE0000} \textbf{ 13.47}} \\
  & CBENet\cite{10RDD2023cvpr} & CVPR/2023 & 22.58 & 0.92 & 24.76 & 25.20 & {\color[HTML]{3166FF} \textbf{ 0.96}} & 17.78 \\
   &\cellcolor[HTML]{EFEFEF} Our Method & \cellcolor[HTML]{EFEFEF} CVPR/2024 &\cellcolor[HTML]{EFEFEF}  {\color[HTML]{FE0000} \textbf{32.54}} & \cellcolor[HTML]{EFEFEF}  {\color[HTML]{FE0000} \textbf{0.96}} & \cellcolor[HTML]{EFEFEF}  {\color[HTML]{FE0000} \textbf{6.28}} & \cellcolor[HTML]{EFEFEF}  {\color[HTML]{FE0000} \textbf{27.28}} & \cellcolor[HTML]{EFEFEF}  {\color[HTML]{FE0000} \textbf{0.97}} & \cellcolor[HTML]{EFEFEF} \textbf{14.96} \\
  \bottomrule
\end{tabular}
 \caption{The quantitative comparison between our method and the SOTA methods on Adobe \cite{39Bako2016}  and FSDSRD \cite{47matsuo2022fsdsrd} datasets. $\uparrow$ means the larger the better while $\downarrow$ means the smaller the better. The best and second best results are marked in red and blue, respectively. }
    \label{tab:dataset1}
\end{center}
\end{table*}

Our shadow-robust feature loss relies on the perceptual features derived from a pre-trained VGG-19 network \cite{08natual2022VGGcolor,32Dcshadownet2021comparative}. We observe that with an increase in the number of convolutional layers in the VGG-19 network, the extracted feature map, as shown in Fig.\ref{fig:loss}, becomes less affected by shadows. The VGG-19 can be divided into five slices, each ending with a MaxPooling layer.

Subsequently, we extract feature maps from these five slices and conduct a weighted summation to achieve a feature representation that is minimally impacted by shadows, as shown in Eq.\ref{con:IfEq}. 



\begin{equation}
  \begin{aligned} 
    {I}^{f} = \displaystyle\sum_{i=0}^{4}({w}_{i}\times {s}_{i})\label{con:IfEq}\\[0.2mm]
    \end{aligned}
 \end{equation} 
 where $s_i$ represents the feature map of the VGG-19 network in the the $k$-th slice ($k$ = 0, 1, ... , 4). $w_i$ represents the corresponding weight each slice. ${I}^{f}$ represents the feature map of shadow image and shadow-free image. 


These feature maps will be used for the computation of our shadow-robust feature loss. $L_{fea}$ is given by Eq.\ref{con:LfeaEq}:

\begin{equation}
  \begin{aligned} 
    {\mathcal{L}}_{fea} = {\mathbb{E}}_{t\sim [1,T]}{\parallel{I}^{f}_{t}-{I}^{f}_{gt}  \parallel }_{F}^{2} \label{con:LfeaEq}\\[0.2mm]
    \end{aligned}
 \end{equation} 
 where  ${I}^{f}_{t}, {I}^{f}_{gt}$ represent the feature map of shadow image at step t and shadow-free image, respectively. 

The total loss function is depicted in Eq.\ref{con:LtotalEq}.

\begin{equation}
  \begin{aligned} 
    {\mathcal{L}}_{total} = \lambda {\mathcal{L}}_{diff}+ (1-\lambda ){\mathcal{L}}_{fea} \label{con:LtotalEq}\\[0.2mm]
    \end{aligned}
 \end{equation} 
 where $\lambda$ represents the weighting coefficient to balance the influence of each term.


\section{Experiments}

\subsection{Experimental Setups}

\textbf{Implementation details.}
The proposed method is implemented using PyTorch 1.8.0 with Python 3.8, which is trained using on NVIDIA A100-PCIE-40GB GPU and CUDA 11.1. For all experiments, we use the same setting of NAFBlock. The batch sizes are set to 4 and the training patches are $128\times128$ pixels. We use the Lion optimizer with ${\beta }_{1} = 0.9$ and ${\beta }_{2} = 0.99$. The initial learning rate is set to $3\times10^{-5}$ and is decayed to $1e^{-7}$ by the Cosine scheduler. The noise level is fixed to 50, the number of diffusion denoising steps is set to 100, set training iterations to 500 000 and set the $\lambda = 0.5$.

\begin{table*}[h]
  \begin{center}
  \begin{tabular}{*{8}{c|c|ccc|ccc}}
  \toprule
  \multirow{2}{*}{\makecell[c]{Methods}} &  \multirow{2}{*}{\makecell[c]{Venue/Year}} & \multicolumn{3}{c|}{\makecell[c]{SDCSRD Datset}} & \multicolumn{3}{c}{\makecell[c]{RDD Dataset}}  \\
  & & \makecell[c]{$PSNR\uparrow$} &  \makecell[c]{$SSIM\uparrow$} &  \makecell[c]{$RMSE\downarrow $} & \makecell[c]{$PSNR\uparrow$} &  \makecell[c]{$SSIM\uparrow$} &  \makecell[c]{$RMSE\downarrow $}  \\
  \midrule
  ISR-based\cite{25iterative2018comparative} & ICASSP/2018 & 23.55 & 0.90 & 21.39 & 22.15 & 0.87 & 21.53 \\
  3D-PC-based\cite{26kligler2018comparative} & CVPR/2018 & 18.43 & 0.87 & 32.54 & 16.82 & 0.76 & 37.77 \\
  WF-based\cite{27jung2018comparative} & ACCV/2018 & 15.78 & 0.88 & 46.83  & 16.28 & 0.83 & 44.04 \\
  BE-based\cite{28wang2019comparative} & ICIP/2019 & 14.01 & 0.80 & 55.60  & 13.12 & 0.60 & 67.82 \\
  LGBC-based\cite{29wang2020comparative} & ICASSP/2020 & 16.18 & 0.86 & 43.53  & 15.84 & 0.82 & 47.09 \\
  JWF-based\cite{01wangze2022} & SMC/2022 & 20.29 & 0.82 & 28.43 & 21.04 & 0.84 & 24.87 \\
  BEATE-based\cite{02liuwenjie2023} & ICASSP/2023 & 21.27 & 0.84 & 25.79 & 20.35 & 0.84 & 26.38 \\
  CBENet\cite{10RDD2023cvpr} & CVPR/2023 & {\color[HTML]{0045FE} \textbf{ 24.58}} & {\color[HTML]{0045FE} \textbf{ 0.93}} & {\color[HTML]{0045FE} \textbf{ 18.56}} & {\color[HTML]{FE0000} \textbf{ 35.15}} & {\color[HTML]{FE0000} \textbf{ 0.96}} & {\color[HTML]{FE0000} \textbf{ 4.69}} \\
  \rowcolor[HTML]{EFEFEF} 
  Our Method & CVPR/2024 & {\color[HTML]{FE0000} \textbf{41.66}} & {\color[HTML]{FE0000} \textbf{0.99}} & {\color[HTML]{FE0000} \textbf{2.64}} & {\color[HTML]{0045FE} \textbf{26.51}} & {\color[HTML]{0045FE} \textbf{0.95}} & {\color[HTML]{0045FE} \textbf{14.92}} \\
  \bottomrule
\end{tabular}
\caption{The quantitative comparison between our method and the SOTA methods on SDCSRD(ours) and RDD \cite{10RDD2023cvpr} datasets. $\uparrow$ means the larger the better while $\downarrow$ means the smaller the better. The best and second best results are marked in red and blue, respectively. }
    \label{tab:dataset2}
\end{center}
\end{table*}

\begin{table*}[h]
  \begin{center}
  \begin{tabular}{*{8}{c|ccc|ccc|c}}
  \toprule
  \makecell[c]{Methods} & \makecell[c]{$PSNR\uparrow$} &  \makecell[c]{$SSIM\uparrow$} &  \makecell[c]{$RMSE\downarrow $} & \makecell[c]{$PSNR_Y\uparrow$} &  \makecell[c]{$SSIM_Y\uparrow$} &  \makecell[c]{$RMSE_Y\downarrow $} &  \makecell[c]{$LPIPS\downarrow $}  \\
  \midrule
  w/o NAFBlock & 23.20 & 0.89 & 18.24  & 24.56 & 0.92 & 15.58 & 0.10 \\
  w/o SMGDM & 30.60 & 0.94 & 7.56 & 32.94 & 0.96 & 6.48 & 0.04 \\
  w/o $L_{fea}$ & 32.21 & 0.95 & 6.63 & 34.28  & 0.97 & 5.57 & 0.03 \\
  Ours (Complete model) & \textbf{32.54} & \textbf{0.96} & \textbf{6.28} & \textbf{34.66} & \textbf{0.98} & \textbf{5.00} & \textbf{0.02} \\
  \bottomrule
\end{tabular}
\caption{ Quantitative comparison of our complete model w/o NAFBlock, w/o SMGDM, and w/o $L_{fea}$ on Adobe dataset respectively. $PSNR_Y$, $SSIM_Y$, and $RMSE_Y$ represent the $PSNR$, $SSIM$ and $RMSE$ in $Y$ channal of the $YCrCb$ color space, respectively. }
    \label{tab:Ablationstudy}
\end{center}
\end{table*}

\textbf{Datasets.}
We carry out experiments on four benchmark datasets for the comparison between our DocShaDiffusion and the SOTA methods: (1) SDCSRD dataset (synthetic dataset) includes 17624 tuples (shadow image, shadow-free image), of which 13200 pairs are allocated for training, 3300 pairs for validation, and 1124 pairs for testing. (2) FSDSRD dataset (synthetic dataset) \cite{47matsuo2022fsdsrd}, consisting of 710 tuples, is employed for testing. (3) Adobe dataset (real dataset) \cite{39Bako2016}, containing 81 tuples, is used for testing. (4) RDD dataset (real dataset) \cite{10RDD2023cvpr}, which includes 545 tuples, is employed for testing.

\textbf{Evaluation measures.}
Following the previous works\cite{30maskgan2019comparative, 31unsupervised2020comparative, 32Dcshadownet2021comparative, 10RDD2023cvpr,06cvpr2023shadowdiffusion}, we utilize the root mean square error (RMSE) in the RGB color space as the quantitative evaluation metric of the shadow removal results, compared with the corresponding ground truth. Besides, we also adopt the Peak Signal-to-Noise Ratio (PSNR) and the structural similarity (SSIM) to measure the performance of various methods in the RGB color space. For the PSNR and SSIM metrics, higher values represent better results. We also measure the $PSNR_Y$, $SSIM_Y$, and $RMSE_Y$, which represent the RMSE, PSNR and SSIM in Y channal of the YCrCb color space, respectively. Finally, we use the learned perceptual image patch similarity (LPIPS) as one of the main evaluation metrics.

\subsection{Comparison with State-of-the-Art}

We compare our network with ten popular or SOTA methods, which include seven heuristic-based methods \cite{25iterative2018comparative, 26kligler2018comparative, 27jung2018comparative, 28wang2019comparative, 29wang2020comparative, 01wangze2022, 02liuwenjie2023} and three neural network-based methods \cite{30maskgan2019comparative, 32Dcshadownet2021comparative, 10RDD2023cvpr}. Among these, CBENet \cite{10RDD2023cvpr} is a document image shadow removal method, while the other two are algorithms for shadow removal in natural scene images. For heuristic learning, we utilize the code provided by the original authors. The results of other models are provided by MS-GAN \cite{30maskgan2019comparative}, DCShadow-Net\cite{32Dcshadownet2021comparative}, CBENet \cite{10RDD2023cvpr}.

\textbf{Quantitative comparison.}
Table \ref{tab:dataset1} presents the PSNR, SSIM, and RMSE on the Adobe \cite{39Bako2016} and FSDSRD\cite{47matsuo2022fsdsrd} datasets, respectively. It can be observed that for the Adobe dataset, our method outperforms the SOTA method across PSNR, SSIM, and RMSE metrics. On FSDSRD dataset, although the RMSE does not achieve the best score, both PSNR and SSIM reach optimal levels, with SSIM notably increasing from 0.95 to 0.97.

Table \ref{tab:dataset2} displays the PSNR, SSIM, and RMSE on the SDCSRD and RDD \cite{10RDD2023cvpr} datasets respectively. Our method significantly leads across all metrics on the SDCSRD dataset when compared to the SOTA method (CBENet). On the RDD test dataset, although our model is trained on the SDCSRD dataset and CBENet is trained on the RDD dataset, our results do not surpass those of CBENet. Nevertheless, our metrics are still superior to other methods. And Table \ref{tab:datasetall} displays that, on average across all datasets, our model significantly outperforms CBENet.

\begin{table}[h]
  \begin{center}
  \begin{tabular}{*{4}{c|ccc}}
  \toprule
  \multirow{2}{*}{\makecell[c]{Methods}} & \multicolumn{3}{c}{\makecell[c]{Average}} \\
  & \makecell[c]{$PSNR\uparrow$} &  \makecell[c]{$SSIM\uparrow$} &  \makecell[c]{$RMSE\downarrow $} \\
  \midrule
  ISR-based\cite{25iterative2018comparative}  & 25.53 & 0.91 & 16.68  \\
  3D-PC-based\cite{26kligler2018comparative}  & 18.73 & 0.84 & 33.14  \\
  WF-based\cite{27jung2018comparative} &  15.90 & 0.86 & 48.78   \\
  BE-based\cite{28wang2019comparative}  & 10.50 & 0.53 & 100.26   \\
  LGBC-based\cite{29wang2020comparative}  & 17.52 & 0.88 & 39.62   \\
  JWF-based\cite{01wangze2022}  & 23.01 & 0.84 & 21.40  \\
  BEATE-based\cite{02liuwenjie2023}  & 23.45 & 0.86 & 20.41  \\
  CBENet\cite{10RDD2023cvpr}  & {\color[HTML]{0045FE} \textbf{ 26.88}} & {\color[HTML]{0045FE} \textbf{ 0.94}} & {\color[HTML]{0045FE} \textbf{ 16.45}} \\
  \rowcolor[HTML]{EFEFEF} 
  Our Method & {\color[HTML]{FE0000} \textbf{32.00}} & {\color[HTML]{FE0000} \textbf{0.97}} & {\color[HTML]{FE0000} \textbf{9.7}}  \\
  \bottomrule
\end{tabular}
\caption{The quantitative comparison of average values across all datasets between our method and the SOTA methods. }
    \label{tab:datasetall}
\end{center}
\end{table}

\textbf{Qualitative comparison.}
Fig.\ref{fig:VisualResults} displays the qualitative results for each method. It is observable that our results closely resemble the GT. Notably, first, our method better ensures the invariance of the background color in document images: as illustrated in rows 2, 3, and 6 of the Fig.\ref{fig:VisualResults}, the results of the SOTA model, CBENet \cite{10RDD2023cvpr}, cause the colored background in the shadowed areas to turn white, whereas our results can accurately restore the original background colors in those areas. Second, our method effectively removes severe shadows: as shown in rows 4 and 5 of the Fig.\ref{fig:VisualResults}, the edges of the shadows are not completely removed in the images generated by CBENet \cite{10RDD2023cvpr}, while our approach successfully eliminates all shadowed regions. Overall, our method achieves superior visual effects compared to others across a variety of scenarios including printed documents, posters, and fonts.

\subsection{Ablation Study}

We have evaluated each module of our network on Adobe datasets\cite{39Bako2016} for ablation study, as shown in Table \ref{tab:Ablationstudy}. Notably, the  w/o NAFBlock in the first row of Table \ref{tab:Ablationstudy}, we use UNet to replace NAFBlock in the encoder module and decoder module.  For the  w/o SMGDM in the second row of Table \ref{tab:Ablationstudy}, we use a fully noise image as $X_T$ for the SMGDM process.  For the  w/o $L_{fea}$ in the third row of Table \ref{tab:Ablationstudy}, we remove this loss computation. 

It can be concluded from Table \ref{tab:Ablationstudy} that the NAFBlock plays important role in improving performance, the PSNR increases from 23.20dB to 32.54dB, SSIM increases from 0.89 to 0.96, and most notably, LPIPS decreases from 0.10 to 0.02.  The  SMGDM is effective to enhance the model's performance, for example, PSNR increases from 30.60 to 32,54. Additionally, the incorporation  of $L_{fea}$  proves beneficial for shadow removal from document images across all evaluation metrics, further enhancing model's performance.

Therefore, it is evident that our DocShaDiffusion method effectively removes shadows from document images across various datasets. In the future, it is expected that our diffusion model can be applied to other similar tasks such as image demoiréing, watermark removal. 

 
\section{Conclusion}

In this paper, we propose the first diffusion-based model in the latent space for document image shadow removal, which is called DocShaDiffusion. We design a shadow mask-aware guided diffusion module (SMGDM), which is supervised using the shadow soft-mask obtained from shadow soft-mask generation module (SSGM). It focuses on shadow portions during both the forward diffusion process and the reverse denoising stages. Experimental results on SDCSRD, Adobe, and FSDSRD datasets shed light on the effectiveness of our model, outperforming the SOTA methods. We propose a synthetic dataset called SDCSRD, which can provide powerful support data support and enrich the research  of document enhancement.

\bibliographystyle{IEEEtran}
\bibliography{refs}

\newpage

 
\vspace{11pt}

\begin{IEEEbiography}[{\includegraphics[width=1in,height=1.25in,clip,keepaspectratio]{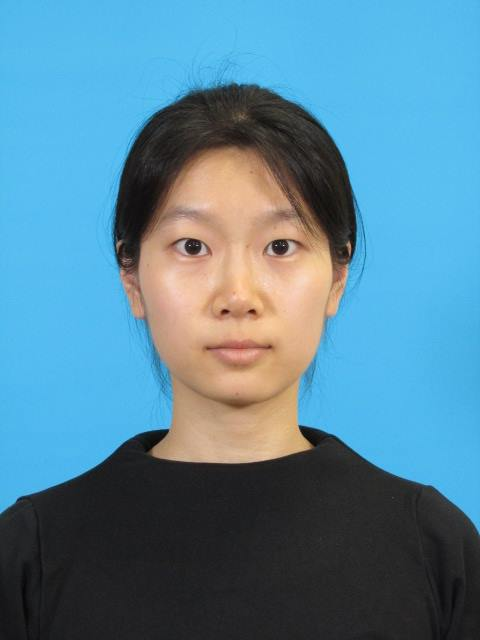}}]{Wenjie Liu}
received her M.S. degree in software engineering from Northwestern Polytechnical University, Xi’an, China, in 2024. She received the B.S. degree in software engineering from Xi’an University of Texhnology, Xi’an, China, in 2021. Her current research interests include computer vision, deep learning, and artificial Intelligence generated content.
\end{IEEEbiography}

\begin{IEEEbiography}[{\includegraphics[width=1in,height=1.25in,clip,keepaspectratio]{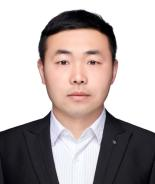}}]{Bingshu Wang}
received his Ph.D. degree in Computer Science from University of Macau, Macau, China, in 2020. He received the M.S. degree in electronic science and technology (Integrated circuit system) from Peking University, Beijing, China, in 2016,  and B.S. degree in computer science and technology from Guizhou University, Guiyang, China, in 2013. Now he is an associate professor in School of Software, Northwestern Polytechnical University. He is also an member of Chinese Association of Automation (CAA), China Computer Federation (CCF), Chinese Association for Artificial Intelligence(CAAI). His current research interests include computer vision, intelligent video analysis and machine learning. 
\end{IEEEbiography}

\begin{IEEEbiography}[{\includegraphics[width=1in,height=1.25in,clip,keepaspectratio]{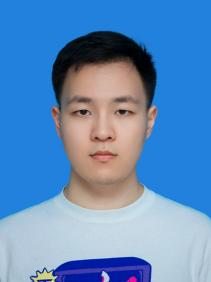}}]{Ze Wang}
received his M.S. degree in software engineering from Northwestern Polytechnical University, Xi’an, China, in 2024. He received the B.S. degree in software engineering from Northeast Forestry University, Harbin, China, in 2021. His current research interests include computer vision, deep learning, and artificial Intelligence generated content.
\end{IEEEbiography}

\begin{IEEEbiography}[{\includegraphics[width=1in,height=1.25in,clip,keepaspectratio]{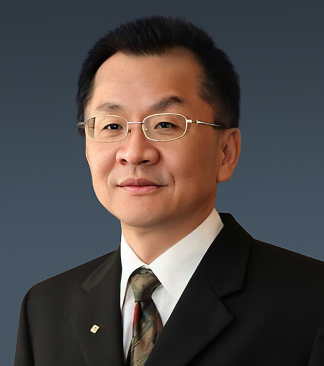}}]{C. L. Philip Chen}
 (S’88–M’88–SM’94–F’07) is the Chair Professor and Dean of the College of Computer Science and Engineering, South China University of Technology.  He is a Fellow of IEEE, AAAS, IAPR, CAA, and HKIE; a member of Academia Europaea (AE), and a member of  European Academy of Sciences and Arts (EASA). He received IEEE Norbert Wiener Award in 2018 for his contribution in systems and cybernetics, and machine learnings, received two times best transactions paper award from IEEE Transactions on Neural Networks and Learning Systems for his papers in 2014 and 2018 and he is a highly cited researcher by Clarivate Analytics from 2018-2022. His current research interests include cybernetics, systems, and computational intelligence. 

He was the Editor-in-Chief of the IEEE Transactions on Cybernetics, the Editor-in-Chief of the IEEE Transactions on Systems, Man, and Cybernetics: Systems, President of IEEE Systems, Man, and Cybernetics Society. Dr. Chen was a recipient of the 2016 Outstanding Electrical and Computer Engineers Award from his alma mater, Purdue University (in 1988), after he graduated from the University of Michigan at Ann Arbor, Ann Arbor, MI, USA in 1985. 
\end{IEEEbiography}



\vfill

\end{document}